\documentclass[conference]{IEEEtran}
\IEEEoverridecommandlockouts
\usepackage{cite}
\usepackage{amsmath,amssymb,amsfonts}
\usepackage{algorithmic}
\usepackage{graphicx}
\usepackage{textcomp}
\usepackage{xcolor}
\usepackage{url}
\usepackage{enumitem}

\usepackage[colorlinks=true, allcolors=blue]{hyperref}

\def\BibTeX{{\rm B\kern-.05em{\sc i\kern-.025em b}\kern-.08em
    T\kern-.1667em\lower.7ex\hbox{E}\kern-.125emX}}
\begin{document}

\title{Optimizing YOLO Architectures for Optimal Road Damage Detection and Classification: A Comparative Study from YOLOv7 to YOLOv10
% \thanks{Authors would like to acknowledge the Institute for Homeland Security at Sam Houston State University to fund this project.}
}

\author{\IEEEauthorblockN{1\textsuperscript{st} Vung Pham}
\IEEEauthorblockA{\textit{Computer Science Department} \\
\textit{Sam Houston State University}\\
Huntsville, Texas, United States \\
vung.pham@shsu.edu}
\and
\IEEEauthorblockN{2\textsuperscript{nd} Lan Dong Thi Ngoc}
\IEEEauthorblockA{\textit{Department of Economic Information System} \\
\textit{Academy of Finance}\\
Hanoi, Viet Nam \\
dongngoclan@hvtc.edu.vn}
\and
\IEEEauthorblockN{3\textsuperscript{rd} Duy-Linh Bui}
\IEEEauthorblockA{\textit{FPT Polytechnic International} \\
\textit{FPT University}\\
Hanoi, Vietnam \\
linhbd2@fpt.edu.vn}
}
\IEEEoverridecommandlockouts
\IEEEpubid{\makebox[\columnwidth]{978-1-6654-8045-1/22/\$31.00~\copyright2022 IEEE\hfill} \hspace{\columnsep}\makebox[\columnwidth]{ }}
\maketitle
\IEEEpubidadjcol
\begin{abstract}
% road damages
% proposed approach
% experiments
% results
Maintaining roadway infrastructure is essential for ensuring a safe, efficient, and sustainable transportation system. However, manual data collection for detecting road damage is time-consuming, labor-intensive, and poses safety risks. Recent advancements in artificial intelligence, particularly deep learning, offer a promising solution for automating this process using road images. This paper presents a comprehensive workflow for road damage detection using deep learning models, focusing on optimizations for inference speed while preserving detection accuracy. Specifically, to accommodate hardware limitations, large images are cropped, and lightweight models are utilized. Additionally, an external pothole dataset is incorporated to enhance the detection of this underrepresented damage class. The proposed approach employs multiple model architectures, including a custom YOLOv7 model with Coordinate Attention layers and a Tiny YOLOv7 model, which are trained and combined to maximize detection performance. The models are further reparameterized to optimize inference efficiency. Experimental results demonstrate that the ensemble of the custom YOLOv7 model with three Coordinate Attention layers and the default Tiny YOLOv7 model achieves an F1 score of 0.7027 with an inference speed of 0.0547 seconds per image. The complete pipeline, including data preprocessing, model training, and inference scripts, is publicly available on the project's GitHub repository, enabling reproducibility and facilitating further research.
\end{abstract}


\section{Introduction}
The condition of roadway infrastructure is critical for ensuring transportation safety, efficiency, and sustainability. Road damages such as cracks, potholes, and surface wear not only degrade the quality of transportation but also pose significant risks to drivers and incur high maintenance costs. Traditional methods of road condition assessment rely on manual inspection, which is time-consuming, labor-intensive, and prone to human error. Moreover, manual data collection exposes inspectors to hazardous environments, further emphasizing the need for automated solutions. Advancements in artificial intelligence (AI) and deep learning offer new opportunities to automate the detection and classification of road damages from image data, promising improved accuracy, speed, and cost-efficiency~\cite{pham2020road, pham2022road}.

Automating road damage detection has the potential to significantly improve maintenance planning and resource allocation. Rapid and accurate damage detection can help transportation agencies prioritize repairs, minimize disruptions, and optimize budgets. Furthermore, timely detection of road damages is crucial for ensuring public safety and preventing minor issues from escalating into major problems that require extensive repairs. Despite these benefits, developing automated detection systems that can handle diverse road conditions across different regions and capture variations in damage types remains a challenging task \cite{pham2020road, pham2022road}. Furthermore, it has become increasingly crucial to address resource optimization concerns, particularly regarding inference speed and memory usage, to enable real-time deployment of these models. Specifically, results from previous IEEE BigData cup challenges in road damage detection show that training large models or ensembling more models often leads to better accuracy~\cite{arya2020global, arya2021deep, arya2022crowdsensing}, but at the same time suffer from inference speed and hardware requirements for deployment~\cite{arya2024global}.

State-of-the-art object detection models such as YOLO (You Only Look Once) have shown promising results in various computer vision tasks~\cite{pham2022handon}, including road damage detection. The YOLOv7 model, in particular, offers a good balance between detection accuracy and computational efficiency~\cite{pham2022road}. However, working with large, high-resolution images poses challenges for both training and inference due to hardware limitations. Additionally, existing road damage datasets often lack sufficient representation of certain damage types, such as potholes, leading to imbalanced detection performance across classes. Therefore, this project employs an optimized YOLOv7-based approach for road damage detection and classification to address these challenges. The primary contributions of this work include the following:

\begin{itemize}
    \item \textbf{Integrated and Balanced Diverse Datasets}: Combined a custom road damage dataset with an external Pothole dataset to address class imbalance and enhance the model's ability to detect underrepresented damage types.
    
    \item \textbf{Developed an Optimized YOLOv7-based Detection Model}: Introduced a custom YOLOv7 model with Coordinate Attention layers and ensembled it with a Tiny YOLOv7 model, achieving improved detection accuracy and inference efficiency.

    \item \textbf{Optimized Inference for Real-time Applications}: Implemented model reparameterization techniques to reduce computational overhead, resulting in a faster inference while maintaining high detection performance.
    
\end{itemize}

\section{Related Work}
\subsection{Popular object detection techniques}
Object detection has been a central focus in computer vision research, with significant advancements in recent years. This review examines key techniques that have shaped the field, including two-stage detectors like Faster R-CNN, single-stage detectors like SSD, and the evolution of the YOLO family. Faster R-CNN, introduced by Ren et al. in 2015~\cite{ren2015faster}, marked a significant improvement in two-stage object detection. It builds upon its predecessors by introducing a Region Proposal Network (RPN) that shares convolutional features with the detection network. This innovation allows for nearly cost-free region proposals, significantly improving both speed and accuracy. Faster R-CNN's end-to-end training capability and real-time detection speeds on GPUs made it a benchmark in object detection research. Detectron2~\cite{wu2019detectron2, pham2022handon}, developed by Facebook AI Research, is not a specific algorithm but a powerful object detection library built on PyTorch. It provides implementations of various state-of-the-art detection algorithms, including Faster R-CNN variants. Detectron2's modular design allows for easy customization and offers high-performance implementations of multiple detection algorithms.

Single Shot Detector (SSD), proposed by Liu et al. in 2016~\cite{liu2016ssd}, is a single-stage detector that aims to balance speed and accuracy. It eliminates the need for region proposal networks by using a set of default bounding boxes with different scales and aspect ratios. SSD's faster inference times compared to two-stage detectors and competitive accuracy, especially for larger objects, made it a popular choice for real-time applications2. Also in the single-stage detector family, the YOLO (You Only Look Once) family has undergone several iterations, each improving upon its predecessors. Recently, YOLOv7~\cite{wang2023yolov7}, introduced in 2022, brought significant improvements in both speed and accuracy. It introduced features like extended and compound scaling of depth and width, planned re-parameterized convolution, and dynamic label assignment4. YOLOv8~\cite{yolov8_2023}, released in 2023, further refined the architecture with a focus on real-time performance and improved accuracy on small objects. YOLOv9~\cite{wang2024yolov9}, one iteration in the YOLO family, introduced in early 2024, builds upon its predecessors with innovative features like Programmable Gradient Information and Implicit Knowledge Learning. These advancements allow YOLOv9 to achieve state-of-the-art performance while maintaining efficiency in terms of computational resources and inference speed.

As of the time of this project, YOLOv10~\cite{wang2024yolov10} has not been officially released, but researchers and developers in the computer vision community anticipate its arrival, expecting further improvements in accuracy, speed, and versatility based on the trajectory of previous YOLO versions. Each iteration of YOLO has progressively improved speed and accuracy, making the YOLO family one of the most popular choices for real-time object detection tasks. The continuous evolution of these techniques demonstrates the rapid pace of advancement in object detection, enabling a wide range of applications in computer vision.

\subsection{Deep learning techniques for road damage detection}

In the field of road damage detection, several state-of-the-art solutions have been proposed to tackle the challenges of detecting and classifying roadway defects. Arya et al. \cite{arya2020global, arya2021deep, arya2022crowdsensing} presented a comprehensive series of methods for global roadway damage detection, emphasizing the development of robust models for diverse road conditions. Pham et al. \cite{pham2020road} explored the use of Detectron2’s Faster R-CNN implementation, highlighting its effectiveness in detecting road damages with high precision. Similarly, other studies have investigated the use of YOLO and ensemble approaches for the same task \cite{hegde2020yet, pham2022road}. The findings from these studies suggest that while Faster R-CNN generally achieves superior detection accuracy, it comes at the cost of increased prediction time compared to the YOLO model. On the other hand, the Single Shot Detector (SSD) model strikes a balance between accuracy and speed, offering an intermediate solution for real-time detection tasks \cite{pham2020road, pham2022road}.

The field of road damage detection is evolving rapidly, necessitating continued experimentation to identify optimal approaches for specific use cases. Among the various methods, the YOLO family of models has gained significant attention due to the frequent release of new versions, each offering improvements in performance and functionality. In this project, both accuracy and inference time are critical factors. As a result, we conducted extensive experiments using multiple recent YOLO versions to identify models that achieve a suitable balance between detection precision and computational efficiency.

\section{Datasets}
\subsection{RDD2022 dataset}
Manually labeling road damage in collected images to create training datasets is highly labor-intensive. Therefore, we utilize existing datasets as benchmarks to train deep learning models for automatically detecting road damage in video data. Among the available datasets, the one provided by Sekilab for the IEEE 2020 Big Data Challenge Cup (RDD2020) is particularly useful~\cite{pham2022road}. The RDD2020 dataset~\cite{arya2021rdd2020} contains images labeled with road damage from three countries: the Czech Republic, India, and Japan.

In the same competition (RDD2020), experiments revealed that deep learning models trained on data from one country did not perform well when applied to data from another. To address this issue, the IEEE 2022 Big Data Challenge Cup (CRDD2022), organized by the same group, adopted a crowd-sensing approach and expanded the dataset to include images from China (captured via drone and motorbike), Norway, and the United States. This new dataset, called RDD2022~\cite{arya2022rdd2022a}, is thoroughly described in their work~\cite{arya2022rdd2022}. Table~\ref{tab:traintestimages} provides an overview of the number of training images, test images, and training labels for each country in the RDD2022 dataset. The countries are listed in order of the number of training labels, which is important for model training. Japan has the largest dataset, with 10,506 training images and 16,470 labels, while China Drone has the smallest, with 2,401 training images and 3,068 labels. Notably, no test data is available for China Drone.

\def\arraystretch{1.3}
\begin{table}[htbp]
\caption{The numbers of train and test images and train labels in the Crowdsensing-based Road Damage Detection Dataset (RDD2022).}
\begin{center}
    \begin{tabular}{|l|r|r|r|}
        \hline
        \textbf{Country} & \textbf{Train images} & \textbf{Test images} & \textbf{Train labels} \\
        \hline
        Japan & 10,506 & 2,627 & 16,470 \\
        \hline
        Norway &  8,161 & 2,040 & 11,229 \\
        \hline
        United States & 4,805 & 1,200 & 11,014 \\
        \hline
        India & 7,706 & 1,959 & 6,831 \\
        \hline
        China Motor & 1,977 & 500 & 4,650\\
        \hline
        China Drone & 2,401 & 0 & 3,068 \\
        \hline
        Czech & 2,829 & 709  & 1,745 \\
        \hline
    \end{tabular}
\label{tab:traintestimages}
\end{center}
\end{table}

\subsection{Pothole dataset}
The RDD2022 dataset categorizes training labels into four types: vertical cracks (D00), traversal cracks (D10), alligator cracks (D20), and potholes (D40). Table~\ref{tab:cracktypes} illustrates the distribution of these crack types within the dataset. Two noteworthy observations emerge from this distribution. First, the number of traversal cracks is significantly lower than that of vertical cracks (68 versus 1,555), indicating a potential need for data validation efforts. Second, with the exception of India, the incidence of pothole cracks is comparatively low in all other countries relative to the other damage types.

\begin{table}[htbp]
\caption{Damage type distribution in the Crowdsensing-based Road Damage Detection Challenge (CRDDC2022).}
\begin{center}
    \begin{tabular}{|l|r|r|r|r|}
        \hline
        \textbf{Country} & \textbf{Vertical} & \textbf{Traversal} & \textbf{Alligator} & \textbf{Pothole} \\
        \hline
        Japan & 4,048 & 3,979 & 6,199 & 2,243 \\
        \hline
        Norway &  8,570 & 1,730 & 468 & 461 \\
        \hline
        United States & 6,750 &  3,295 & 834 & 135 \\
        \hline
        India & 1,555 & 68 & 2,021 & 3,187\\
        \hline
        China Motor & 2,678  & 1,096  & 641 & 235 \\
        \hline
        China Drone & 1,426 & 1,263  & 293 & 86 \\
        \hline
        Czech &  988 & 399 & 161 & 197 \\
        \hline
    \end{tabular}
\label{tab:cracktypes}
\end{center}
\end{table}

\begin{figure*}[!htb]
    \centering
    \includegraphics[width=0.8\linewidth]{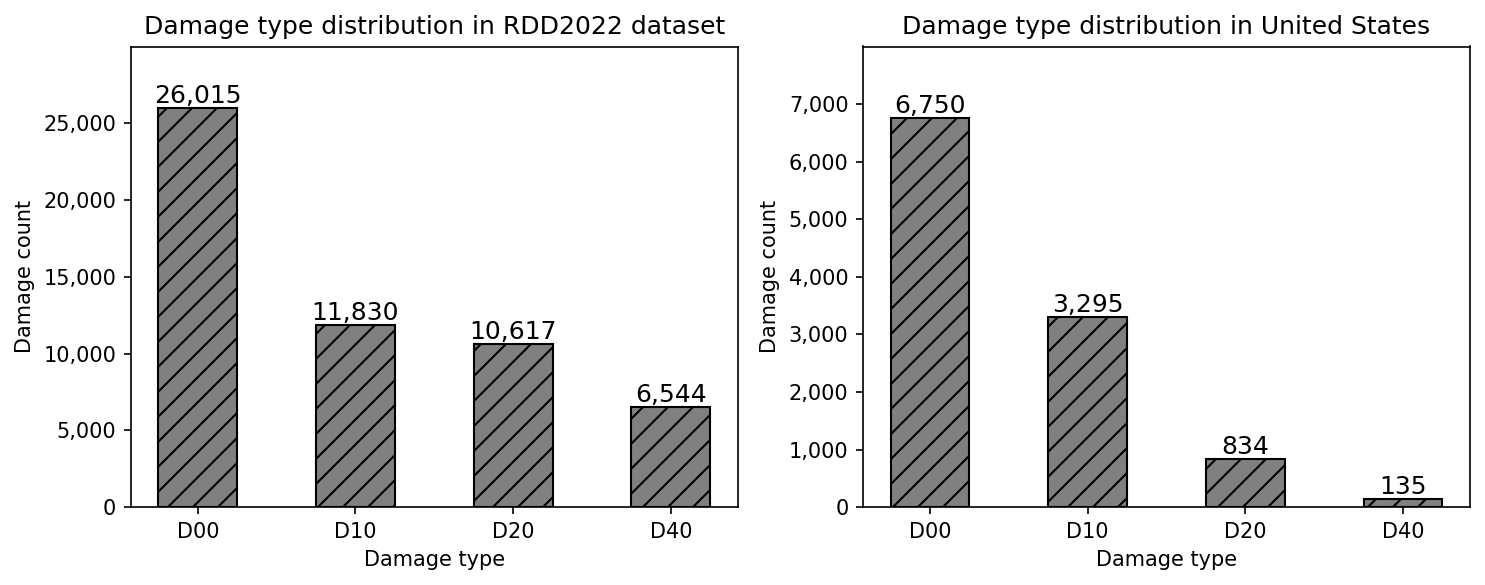}
    \caption{Distribution of damage types in the RDD2022 dataset.}
    \label{fig:damagedistribution}
\end{figure*}

The issue of class imbalance is particularly pronounced in developed countries, where pothole damages are considered more severe and are often repaired promptly, resulting in fewer recorded occurrences. Visualizing the distribution of crack types can help highlight this problem. Figure~\ref{fig:damagedistribution} (left) illustrates the number of pothole damages in the RDD2022 dataset, clearly indicating that the D40 (pothole) category has the fewest occurrences. This imbalance is even more pronounced in the United States, as depicted in Figure~\ref{fig:damagedistribution} (right), which shows only 135 potholes among a total of 11,014 damages. Consequently, the classification and detection accuracy for the D40 class may be lower than for other classes, as models typically struggle to perform well on underrepresented categories.

To mitigate this class imbalance, we incorporate the publicly available Pothole dataset from Roboflow\footnote{\url{https://universe.roboflow.com/brad-dwyer/pothole-voxrl}}. This dataset consists of 665 road images with labeled potholes and is available in various formats compatible with common machine-learning models. For our project, we utilize the YOLOv7 annotation format. While the Pothole dataset contains a single class for potholes identified by the number 0, our dataset labels the pothole class as class 3. Thus, we will update the class identification numbers to ensure consistency across the datasets.

\section{Experiments}
\subsection{Data processing}
\label{sec:dataprocessing}
\subsubsection{Cropping large images}
\label{sec:croppinglargeimages}

Table~\ref{tab:imagesizes} presents an overview of the image sizes used across different folders in the CRDDC2022 dataset. The dimensions of the images vary significantly between countries, both in size and aspect ratio. For instance, images from China Drone and China Motor are formatted at $512 \times 512$ pixels, while those from the Czech Republic are $600 \times 600$ pixels, and images from India are $720 \times 720$ pixels. The folder for Japan contains images of various sizes, including $600 \times 600$, $1,024 \times 1,024$, and $1,080 \times 1,080$ pixels. Norway's folder contains particularly large images, such as $4,040 \times 2,041$ pixels, whereas images in the United States folder have the standard YOLO training image size of $640 \times 640$ pixels.

Training YOLO models on images with varying sizes and aspect ratios can significantly affect both model performance and hardware requirements. Larger images, like those from Norway, demand more GPU memory and computational power, which can lead to slower training processes and increased inference times. Although these larger images can capture more detailed features, they pose challenges when scaled down to standard training sizes (e.g., $640 \times 640$). Specifically, annotations may become too small relative to the image dimensions, and the aspect ratios of the labels may be distorted. This distortion can result in bounding boxes that are too small or even omitted during training, which degrades detection performance for small objects or damages.

For example, an image with dimensions $4,040 \times 2,041$ and a label size of $50 \times 30$ for a vertical crack would scale down to $8 \times 9$ pixels when resized to the standard training size of $640 \times 640$. This transformation introduces two main issues. First, the label size becomes too small, which may cause the thin crack line to be erased, making detection difficult. Second, the change in aspect ratio can alter the label's appearance, causing the vertical crack to look more like a traverse crack instead. Therefore, using standard-sized images (e.g., $640 \times 640$) is generally recommended to balance performance and hardware efficiency while minimizing the risk of losing small annotations. Larger image sizes should only be employed when the hardware supports them or when scaling strategies are implemented to maintain the relative size of annotations in downscaled images.

\begin{table}[htbp]
    \caption{Image sizes for different image folders in the Crowdsensing-based Road Damage Detection Challenge (CRDDC2022).}
    \begin{center}
    \begin{tabular}{|l|r|r|}
        \hline
        \textbf{Folder} & \textbf{Image width(s)} & \textbf{Image height(s)} \\
        \hline
        China Drone & {512} & {512}\\
        \hline
        China Motor & {512} & {512}\\
        \hline
        Czech & {600} & {600}\\
        \hline
        India & {720} & {720}\\
        \hline
        Japan &{600, 1024, 1080, 540} & {600, 1024, 1080, 540}\\
        \hline
        Norway & {4040, 3650, 3643} & {2041, 2035, 2044}\\
        \hline
        United States & {640} & {640} \\
        \hline
        \end{tabular}
    \label{tab:imagesizes}
    \end{center}
\end{table}

\begin{figure}[!htb]
    \centering
    \includegraphics[width=\linewidth]{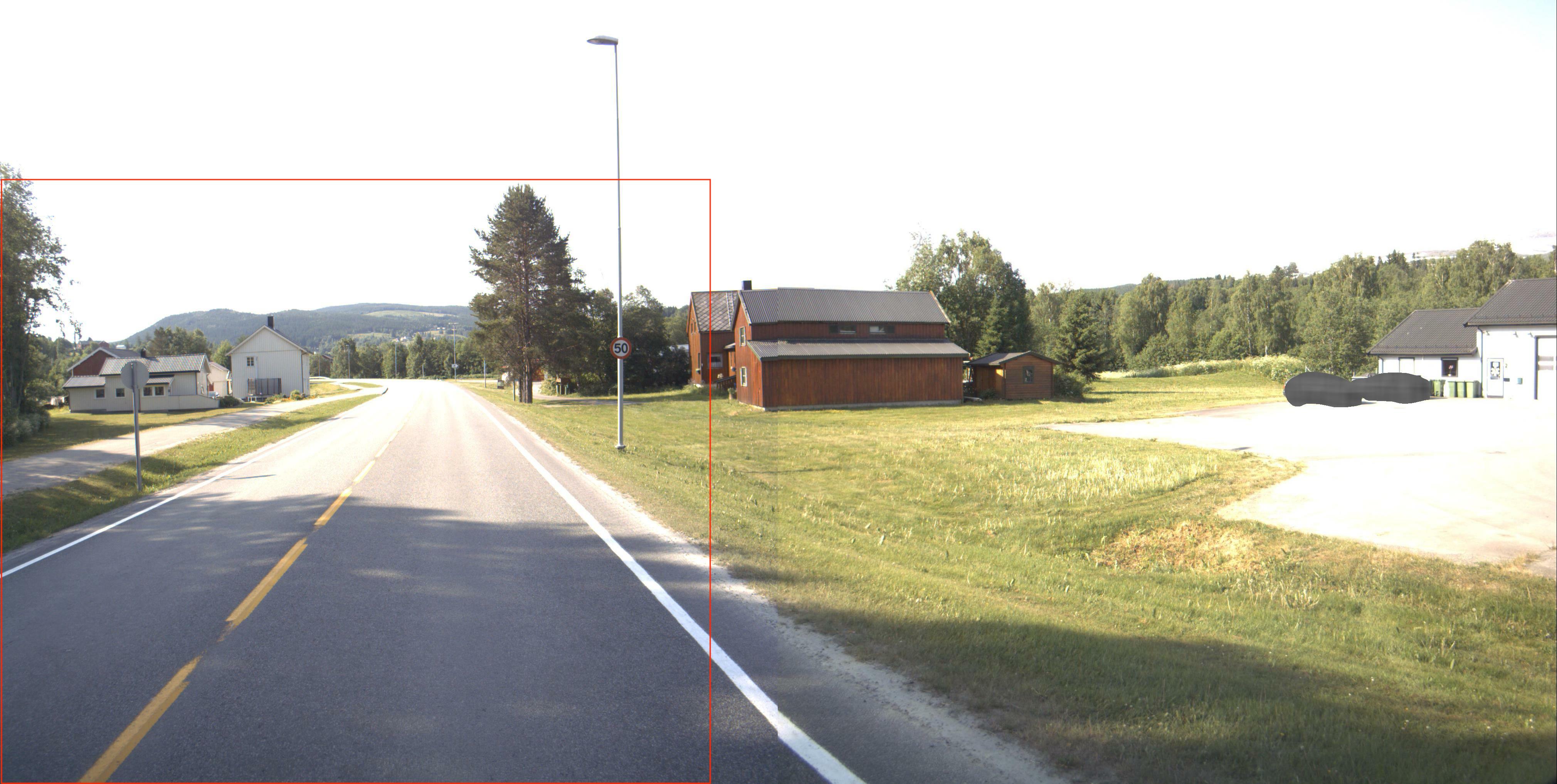}
    \caption{An example image from the Norway folder (\textit{Norway\_000003.jpg}). This large image appears to be a composite of two smaller images tiled from left to right. The majority of the road section and visible damages are concentrated in the lower-left corner, highlighted by the red rectangle with dimensions of $1,824 \times 1,824$ pixels.}
    \label{fig:norwayimage}
\end{figure}

A detailed examination of the images in the Norway folder revealed several notable characteristics. Many of these images appear to be composed of two stitched sections, with one section on the left and the other on the right. Figure~\ref{fig:norwayimage} illustrates this with \textit{Norway\_000003.jpg}, which shows that most of the road areas are concentrated in the lower-left corner of the images. Consequently, we decided to crop each image in the Norway folder to focus on a $1824\times1824$ pixel region in that corner. This resolution was selected because it is approximately half the width of the original images, effectively covering the primary road area, and it meets the YOLOv7 requirement for training images to have dimensions that are multiples of 32. Additionally, we adjusted or removed annotations for any damages outside of this cropped region. The number of discarded or modified annotations was minimal, totaling 537 out of 11,229 road damage instances, suggesting that the selected crop region is appropriate for this dataset.

To streamline the training process, we cropped the original images in the Norway training dataset and adjusted the corresponding annotations to align with the new image sizes and locations. The modified images and annotations were saved in a new folder titled \textit{Norway1}, which replaced the original images as the training data for Norway. For the testing dataset, we customized the default YOLOv7 data loader by implementing two separate loaders: one that processes images with dimensions $width > 1,824$ and $height > 1,824$, and another for images with smaller dimensions. The first loader crops these larger images before making predictions and subsequently readjusts the predicted bounding box locations to correspond with the original image dimensions, ensuring accurate results.

\subsubsection{Annotation conversion and train/validation splitting}
Besides the images with valid annotations, the RDD2022 dataset contains images that either lack annotations or have annotations not corresponding to any of the four damage types considered in this project (D00, D10, D20, and D40). The deep learning technique employed here is YOLOv7, so we converted the existing annotations from XML files in PASCAL VOC format to $*.txt$ files compatible with the YOLOv7 format and remove those that do not have annotations or the annotations are not considered by this project.

To select the best models, we split the training images into training and validation sets. Although it is possible to configure the test size in YOLOv7, we chose to perform this split only once for reproducibility and to facilitate performance comparisons across experiments. Among the seven data folders, the China\_Drone folder does not have a designated test set; therefore, all its images are included in the training set. For the other folders, training images are divided into 90\% for training and 10\% for validation, which is used for selecting the best model.

\subsection{Deep Learning Models}
\subsubsection{Comparing different YOLO versions}
In this project, we explored various YOLO versions, beginning with YOLOv7 (released in 2022) due to its strong performance in prior research~\cite{pham2022road}. YOLOv8 represents a significant advancement in the YOLO object detection framework, offering notable enhancements over YOLOv7. It provides faster detection speeds and the capability to process more frames per second, along with improved accuracy reflected in higher mean average precision (MAP) scores. The anchor-free architecture of YOLOv8 facilitates better adaptation to diverse object sizes and proportions, making it especially effective for detecting small objects. These improvements position YOLOv8 as a powerful tool for applications that demand both precision and rapidity in object detection.

The authors of YOLOv9 have introduced further enhancements over earlier iterations; however, the strengths of YOLOv8 and YOLOv9 may vary depending on the specific use case. YOLOv8 may excel in close-range object detection scenarios, while YOLOv9 might perform better on large, zoomed-out images. Nonetheless, comparative analyses between these two versions remain limited, and their effectiveness can differ based on the particular application and dataset.

YOLOv10, the latest iteration in the YOLO series, brings additional architectural changes. It is reported to be significantly faster than YOLOv8, potentially achieving speeds 1.5 to 2 times greater, and it features a smaller model size with fewer parameters. However, YOLOv8 still holds an advantage in detecting smaller objects. YOLOv10 may require a lower confidence threshold to match the detection performance of YOLOv8. It is important to recognize that while newer versions typically offer improvements, they may also entail trade-offs. For instance, YOLOv10's increased speed may come at the expense of some accuracy, particularly for smaller objects.

When reporting accuracy and performance metrics, these versions are evaluated based on benchmark datasets such as MS COCO. However, it is essential to consider the specific requirements of each use case, the nature of the objects to be detected, and the optimal balance between speed and accuracy when selecting among these versions. Therefore, we conducted experiments with all of these recent YOLO iterations for the specific task and dataset in this project.

\subsubsection{Incorporating Coordinate Attention layers}
\label{sec:coordinateattentionlayers}
YOLOv7 back-end networks frequently utilize channel attention mechanisms to assign different weights to channels during the fusion of features across layers. However, conventional channel attention compresses all spatial information within each channel into a single scalar value, leading to a loss of spatial context. In the RDD2022 dataset, most road damages are located in the lower half of the images, and the appearance of these damages varies depending on their spatial location due to changes in camera perspectives. Therefore, besides training a YOLOv7 model with default configuration, this study also explores the use of Coordinate Attention~\cite{hou2021coordinate}, which captures both channel and positional information more effectively.

\subsubsection{Training hyperparameters}
\label{sec:traininghyperparameters}
Table~\ref{tab:imageaugmentations} shows the final list of image augmentation parameters used in this project (other default parameters remain the same). Specifically, scale, mosaic, mixup, and paste\_in are slightly reduced to avoid unwanted, unrealistic effects. Additionally, road damages collected using cameras placed on car dashboards have some perspective. Thus, we utilized shear (0.01) and perspective (0.0001) in the image augmentation parameters. The spatial information is important in road damage detection tasks. 

\begin{table}[htbp]
    \caption{Experimented image augmentation parameters (other parameters remain the same).}
    \begin{center}
    \begin{tabular}{|l|r|l|}
        \hline
        \textbf{Parameter} & \textbf{Value} & \textbf{Descriptions} \\
        \hline
        scale &  0.7  &   scale (+/- gain) \\
        \hline
        shear&  0.01  &   shear (+/- deg) \\
        \hline
        perspective &  0.0001  &   perspective (+/- fraction) \\
        \hline
        mosaic &  0.5  &   mosaic (probability) \\
        \hline
        mixup &  0.1  &   mixup (probability) \\
        \hline
        paste\_in &  0.05  &   copy paste (probability)\\
        \hline
        \end{tabular}
    \label{tab:imageaugmentations}
    \end{center}
\end{table}

Additionally, for various reasons, the annotated labels are not 100\% accurate or consistent. For instance, many traverse cracks with different perspectives or a slight camera rotation may be mislabeled as longitudinal cracks and vice versa. Additionally, alligator cracks are often confused with longitudinal and traverse cracks. Therefore, we experimented with label smoothing~\cite{szegedy2016rethinking}. Specifically, we apply the label smoothing technique with a rate of 0.1 during training. Finally, we also applied test time augmentation techniques to improve prediction accuracy.

\subsubsection{A tiny YOLOv7 model}
Ensembling models is a well-known strategy for improving prediction accuracy in road damage detections proven by many works in the past~\cite{arya2020global, arya2022crowdsensing}, as it leverages the strengths of multiple models to make more robust decisions. However, this approach typically comes with a trade-off in terms of increased inference time and computational cost. To address these challenges, it is crucial to find a balance between accuracy and efficiency, particularly for real-time applications where latency is a critical concern.

To mitigate the inference time trade-off, we incorporate a smaller variant of YOLOv7 (the Tiny YOLOv7 model) into the ensemble. This allows us to achieve faster predictions without significantly compromising accuracy. In our implementation, we apply the ensemble model only to images of regular size, ensuring a high level of accuracy where needed. For larger images, which inherently require more computation, we rely solely on the standard YOLOv7 model (not ensembling). This selective use of the ensemble ensures that we maintain an optimal balance between inference speed and performance across different image sizes.

\subsection{Inference-time optimization}
\subsubsection{Batch processing}
Batch-based inference can significantly speed up inference time by processing multiple inputs simultaneously, leveraging the parallel processing capabilities of modern hardware like GPUs. This approach allows for more efficient use of computational resources, as many operations can be parallelized across the batch. For YOLO models, especially those with large numbers of parameters, batch processing can lead to substantial speedups by amortizing the overhead of model initialization and data transfer across multiple inputs. However, the effectiveness of batching depends on factors such as input similarity, hardware capabilities, and model architecture. While batching generally improves throughput, it may not always reduce latency for individual requests, making it more suitable for scenarios where processing a large number of inputs is prioritized over minimizing the response time for a single input. Therefore, we also experimented with different batch sizes in this project.

\subsubsection{Tracing}
PyTorch in general and YOLO specifically offer tracing options to convert trained models to a traced version using PyTorch's $torch.jit.trace$ functionality. While tracing typically optimizes the model for faster inference, it may result in lesser flexibility, especially in handling dynamic input sizes. The untraced model ensures consistent behavior with the original version, which can be important for avoiding potential model differentials and mitigating security risks associated with tracing. Additionally, using a traced model may produce specific runtime errors related to invalid image sizes. Although traced models generally offer inference-time benefits, the no-trace option provides more control over the model's behavior and can be preferable in scenarios requiring greater flexibility or when dealing with complex model architectures that don't trace well~\cite{bowen2024night}. Therefore, we also experimented with this option to see if it could help balance between the inference time and accuracy.

\subsubsection{Reparameterization}
Model reparameterization is a technique that restructures the network architecture to enhance inference efficiency without altering the model's behavior. It works by merging multiple convolutional layers and batch normalization operations into a single equivalent convolutional layer during inference. This reduces computational overhead and memory usage, leading to faster inference speeds while maintaining detection accuracy. Some YOLO versions (such as YOLOv7) offer a carefully planned reparameterization strategy that optimizes performance by considering the gradient propagation paths across different network architectures. At the end of the training process, we added code to execute the reparameterization process and saved the reparameterized models for inferencing.

\section{Results and Discussions}
\subsection{Evaluation Metrics and Test Environments}
Different experiments produce different models, so it's important to use a reliable metric to select the best models. Two commonly used evaluation metrics in this context are the Mean Average Precision (mAP) at an Intersection over Union (IoU) threshold of 0.5 (mAP@0.5) and the F1 score. The mAP is useful when assessing a model's stability across varying confidence thresholds, as it provides a more robust evaluation. In contrast, the F1 score is calculated at a specific confidence threshold. A common approach is to use mAP@0.5 on the validation set to select the best model while using the F1 score to report performance on the test set. This project follows that approach: mAP@0.5 is used to select models during training based on validation data, and the F1 score is used to evaluate performance on the test set.

In addition to the F1 score, run time is a critical factor. Both F1 scores and run times are evaluated on the test set by the organizers of the IEEE BigData 2024 Challenge. For faster experimentation, test run times (in seconds) were measured by evaluating the trained models on a local server. The server specifications are 62.5 GiB of memory, an Intel® Xeon(R) Silver 4214R CPU @ 2.40GHz with 48 cores, a 16 GiB GPU, and Ubuntu 20.04.6 LTS as the operating system.

\subsection{Different YOLO Architectures}

Table~\ref{tab:yoloarchitectures} presents the evaluation results for various YOLO architectures. For all these models, we leveraged pre-trained versions to accelerate training convergence through transfer learning. The training and evaluation datasets used for these models are consistent, as described in Section~\ref{sec:dataprocessing}. Each model was trained for 150 epochs with a batch size of 16.

Model 1 refers to the default YOLOv7 architecture, as defined in the \textit{cfg/yolov7.yaml} file. Model 2 is a customized version of YOLOv7, which incorporates three additional Coordinate Attention layers (discussed in Section~\ref{sec:coordinateattentionlayers}) and optimized training hyperparameters (outlined in Section~\ref{sec:traininghyperparameters}). Model 3 is the default YOLOv7 tiny variant, as specified in the \textit{cfg/yolov7-tiny.yaml} file. For the more recent YOLO versions (v8, v9, and v10), we utilized the \textit{ultralytics} package for training. After conducting several experiments, we settled on the configurations \textit{yolov8l}, \textit{yolov9c}, and \textit{yolov10l} for YOLOv8, YOLOv9, and YOLOv10, respectively.

To measure inference time, we performed tests on the evaluation dataset using a batch size of 1 to calculate the per-image inference time for the YOLOv7 models. However, for YOLOv8, v9, and v10, the \textit{ultralytics} package provides the inference time by default, so no additional testing was required for these models.

\begin{table}[htbp]
    \caption{Evaluation results with different YOLO architectures on the evaluation dataset.}
    \begin{center}
    \begin{tabular}{|c|l|r|r|r|r|}
        \hline
        \textbf{Model} & \textbf{Architecture} & \textbf{mAP@0.5} & \textbf{F1} & \textbf{Conf.} & \textbf{Time}\\
        \hline
        1 & YOLOv7 default & 0.661 & 0.63 & 0.231 &  10.4 \\
        \hline
        2 & YOLOv7 custom & 0.662 & \textbf{0.65} & \textbf{0.403} & 15.7 \\
        \hline
        3 & YOLOv7 tiny & 0.616 & 0.61 & 0.244 & \textbf{7.3} \\
        \hline
        4 & YOLOv8 (yolov8l) & \textbf{0.663} & 0.64 & 0.258 & 8.6 \\
        \hline
        5 & YOLOv9 (yolov9c) & \textbf{0.663} & 0.64 & 0.263 & 8.5 \\
        \hline
        6 & YOLOv10 (yolov10l) & 0.647 & 0.62 & 0.238 & 15.6 \\
        \hline
        \multicolumn{6}{l}{Time is in milliseconds per image, Conf.: Confidence} \\
        
        \end{tabular}
    \label{tab:yoloarchitectures}
    \end{center}
    
\end{table}

Table~\ref{tab:yoloarchitectures} presents the experimental results, with the best value for each measurement metric highlighted in bold. YOLOv8 and YOLOv9 (models 4 and 5) show slightly higher mAP@0.5 and F1 scores, along with shorter inference times, compared to the default YOLOv7 model. However, the optimized YOLOv7 model—enhanced with additional Coordinate Attention layers and customized hyperparameters—achieves a better F1 score and confidence threshold. Notably, these same optimizations were also applied to YOLOv8, YOLOv9, and YOLOv10 but did not yield improvements; in some cases, they even degraded performance. As expected, the YOLOv7 tiny version has the lowest accuracy and fastest inference time due to its smaller model size. 

\subsection{Impacts of Inference-Time Optimization Techniques}
\subsubsection{Impacts of batch sizes}
As described in Section~\ref{sec:croppinglargeimages}, we use two separate data loaders for two types of images: large and normal-sized. Using large batch sizes for the larger images could lead to performance bottlenecks, so we consistently set a smaller batch size for these images. Specifically, we tested batch size pairs of (12, 24), (16, 32), and (32, 64), where the first number in each pair refers to the batch size for large images and the second number for normal images. The model used for these experiments was an ensemble of models 1 and 2 from Table~\ref{tab:yoloarchitectures}, with a confidence threshold set at 0.5. The F1 score achieved was 0.7125, and the execution times were 489, 489, and 495 seconds, respectively. These results show that larger batch sizes do not significantly improve performance and can even increase inference time if too large. Therefore, we selected a batch size of (12, 24) for all future experiments.

\subsubsection{Impacts of tracing and reparameterization}

Model reparameterization for the YOLOv7 models used in these experiments does not significantly improve inference time, likely because the selected models are already relatively small. The reparameterization occurs only in the \textit{IDetect} block, where its layers are replaced by fewer layers in the \textit{Detect} block. However, this process has no negative impact on accuracy, so we consistently apply this option in all models.

In our tracing experiments, we first tested an ensemble of two YOLOv7 models (models 1 and 2 from Table~\ref{tab:yoloarchitectures}) with a confidence threshold of 0.5, using a batch size of 12 for large images and 24 for small images. This experiment reduced inference time from 489 to 360 seconds (approximately a 26\% decrease), but it also lowered the F1 score from 0.7125 to 0.6700. We then tested an ensemble of three YOLOv7 models (models 1, 2, and 3) based on the hypothesis that adding more models to the ensemble would improve accuracy while maintaining low inference time through tracing. While this did increase the F1 score to 0.684, the inference time rose to 415 seconds. Therefore, tracing did not yield the expected benefits, as increasing the ensemble size did not maintain a favorable balance between accuracy and inference time.

\subsection{Final Model and Future Work}
For this project, using ensembles of large models improves accuracy but also increases inference time, particularly for large images. Since large images ($width < 1,824$ and $height < 1,824$) require different handling from normal-sized images, we opted for a hybrid approach. Specifically, we used an ensemble of one large model (customized YOLOv7, or model 1 in Table~\ref{tab:yoloarchitectures}) and a tiny YOLOv7 model (model 3) for normal-sized images while only deploying the large model (model 1) for large images. Batch sizes were set to 12 for normal images and 24 for large images. Additionally, the confidence threshold was set to 0.50 for normal images and 0.35 for large images. This configuration resulted in an F1 score of 0.7027 and an inference speed of 0.0547 seconds per image, securing the $4^{th}$ place out of 39 teams worldwide in the Optimized Road Damage Detection Challenge (ORDDC'2024), a track in the IEEE BigData 2024 Challenge.

Table~\ref{tab:yoloarchitectures} indicates that YOLOv8 and YOLOv9 are strong candidates for ensembling, as they exhibit slightly lower F1 scores but better mAP@0.5 and good inference speed. However, the optimizations explored so far have not led to significant improvements in F1 scores. In future work, we plan to explore different optimizations and customizations to enhance the F1 scores. Due to time constraints, we were unable to implement test-time augmentations or experiment with ensembling techniques for the YOLOv8 and YOLOv9 models, which will be a priority moving forward.

\section{Implementation}
The official implementation of YOLOv7 is available at \url{https://github.com/WongKinYiu/yolov7}. However, we have made several modifications to the codebase, including enhancements to the training process, hyperparameters, and data loaders, to improve both accuracy and inference time. For this project, please utilize the customized version of YOLOv7 available in our GitHub repository at \url{https://github.com/phamvanvung/roaddamagedetection2024}, rather than the original implementation. Additionally, for the more recent YOLO versions (v8, v9, and v10), we employed the $ultralytics$ package for training and validation, which can be accessed at \url{https://www.ultralytics.com}.

\section{Conclusion}
This study demonstrates the effectiveness of optimizing YOLO architectures for road damage detection, particularly in balancing detection accuracy with inference speed. By leveraging an ensemble approach combining a customized YOLOv7 model with additional Coordinate Attention layers and a Tiny YOLOv7 model, we achieved competitive results in the IEEE BigData 2024 Challenge, placing $4^{th}$ out of 39 teams. Our approach resulted in an F1 score of 0.7027 and an inference time of 0.0547 seconds per image, showcasing the potential for real-time deployment in resource-constrained environments. Overall, this research highlights the potential of YOLO-based models for practical applications in road damage detection, offering a scalable, efficient solution for maintaining roadway infrastructure.

While our experiments with YOLOv8 and YOLOv9 models showed promise, particularly in terms of mAP@0.5 and inference speed, the optimizations did not yield substantial improvements in F1 scores. Future work will focus on exploring more advanced optimization techniques, including test-time augmentations and additional ensembling strategies, to further improve detection performance. We also plan to investigate the integration of newer YOLO versions and customizations to enhance detection accuracy without compromising speed, particularly for large, high-resolution images.

\section{Author Contributions}
\begin{itemize}[align=left, noitemsep, topsep=0.5em, leftmargin=*]
    \item Vung Pham conceived and designed the research method.
    \item Vung Pham performed the data curation and preprocessing.
    \item Lan Dong and Linh Bui implemented the experiments.
    \item Lan Dong and Linh Bui analyzed the experimental results.
    \item Lan Dong and Linh Bui wrote the original draft of the paper.
    \item Vung Pham reviewed and edited the manuscript.
    \item Vung Pham supervised the project.
    \item All authors have read and approved the final manuscript.
\end{itemize}

\bibliographystyle{./IEEEtran}
\bibliography{IEEEabrv,IEEEfull}

\end{document}